\documentclass{article}

\usepackage{subcaption}
\usepackage{microtype}
\usepackage{graphicx}
\usepackage{booktabs} 

\usepackage{enumitem} 
\usepackage{array} 
\usepackage{xcolor}         
\usepackage{multirow}
\usepackage{color}
\usepackage{colortbl}
\usepackage{makecell}
\usepackage{bbm}

\usepackage{wrapfig}

\usepackage{listings}
\usepackage{xcolor}

\usepackage{hyperref}

\newcommand{\SYSNAME}{\textsc{SteerFair}}

\usepackage[accepted]{icml2024}

\usepackage{amsmath}
\usepackage{amssymb}
\usepackage{mathtools}
\usepackage{amsthm}
\usepackage{enumitem} 
\usepackage[symbol]{footmisc}
\usepackage[capitalize,noabbrev]{cleveref}
\usepackage[export]{adjustbox}

\theoremstyle{plain}

\theoremstyle{definition}

\theoremstyle{remark}

\usepackage{soul}
\definecolor{backcolour}{rgb}{0.95,0.95,0.92}
\lstdefinestyle{mystyle}{
  backgroundcolor=\color{backcolour}, 
  basicstyle=\ttfamily\footnotesize,
  breakatwhitespace=false,         
  breaklines=true,                 
  captionpos=b,                    
  keepspaces=true,                      
  numbersep=5pt,                  
  showspaces=false,                
  showstringspaces=false,
  showtabs=false,                  
  tabsize=2
}

\usepackage[textsize=tiny]{todonotes}
\usepackage{xcolor}

\usepackage{amsmath}

\icmltitlerunning{Discovering Bias in Latent Space: An Unsupervised Debiasing Approach}

\begin{document}

\twocolumn[
\icmltitle{Discovering Bias in Latent Space: An Unsupervised Debiasing Approach}



\icmlsetsymbol{equal}{*}

\begin{icmlauthorlist}
\icmlauthor{Dyah Adila}{equal,sch}
\icmlauthor{Shuai Zhang}{comp}
\icmlauthor{Boran Han}{comp}
\icmlauthor{Yuyang Wang}{comp}
\end{icmlauthorlist}

\icmlaffiliation{comp}{Amazon Web Services}
\icmlaffiliation{sch}{Department of Computer Science, University of Wisconsin-Madison}

\icmlcorrespondingauthor{Dyah Adila}{adila@wisc.edu}
\icmlcorrespondingauthor{Shuai Zhang}{shuaizs@amazon.com}

\icmlkeywords{Machine Learning, ICML}

\vskip 0.3in
]



\printAffiliationsAndNotice{Work done during internship at Amazon Web Services} 

\begin{abstract}
The question-answering (QA) capabilities of foundation models are highly sensitive to prompt variations, rendering their performance susceptible to superficial, non-meaning-altering changes. This vulnerability often stems from the model's preference or bias towards specific input characteristics, such as option position or superficial image features in multi-modal settings. We propose to rectify this bias \textit{directly in the model's internal representation}. Our approach, $\SYSNAME$, finds the bias direction in the model's representation space and steers activation values away from it during inference. Specifically, we exploit the observation that bias often adheres to simple association rules, such as the spurious association between the first option and correctness likelihood. Next, we construct demonstrations of these rules from unlabeled samples and use them to identify the bias directions. We empirically show that $\SYSNAME$ significantly reduces instruction-tuned model performance variance across prompt modifications on three benchmark tasks.
Remarkably, our approach surpasses a supervised baseline with 100 labels by an average of 10.86\% accuracy points and 12.95 score points and matches the performance with 500 labels. 
\end{abstract}

\section{Introduction}
\label{sec_intro}


\begin{figure}[t]
    \centering
    \includegraphics[width=.5\textwidth]{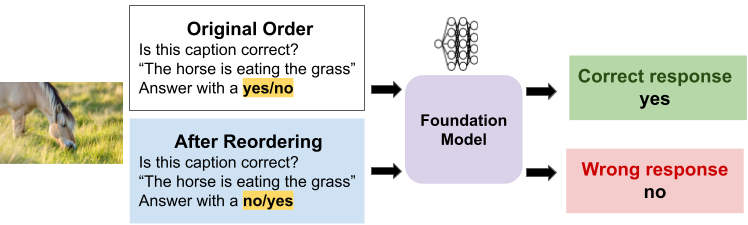}
    \small
    \setlength\tabcolsep{4pt}
     \vspace{1em}
    \begin{tabular}{lccc|cc}
        \toprule
        \multirow{2}{*}{Model} & \multicolumn{3}{c}{ScienceQA (2 options)} & \multicolumn{2}{c}{VGR} \\ 
        \cmidrule(lr){2-4}  \cmidrule(lr){5-6} 
        & Original & A & B & yes/no & no/yes   \\
        \toprule
        LLaVA & 64.99\% & \cellcolor{green!10}{68.22\%} & \cellcolor{red!10}{63.15\%} & \cellcolor{green!10}{82.29\%} & \cellcolor{red!10}{66.10\%}  \\
        IDEFICS & 61.04\% & \cellcolor{green!10}{85.10\%} &\cellcolor{red!10}{34.11\%} & \cellcolor{green!10}{40.71\%} & \cellcolor{red!10}{61.26\%}  \\
        InstructBLIP & 61.27\% & \cellcolor{green!10}{83.57\%} & \cellcolor{red!10}{37.66\%} & \cellcolor{green!10}{56.27\%} & \cellcolor{red!10}{22.01\%} \\
        \bottomrule
    \end{tabular}
    \caption{Top: Model predictions are sensitive to prompt order changes. Bottom: Performance of instruction-tuned models on (1) ScienceQA (2 options) in the original order and with golden answers moved to A/B, and (2) Visual Genome Relation (VGR) with prompt variations using "yes/no" and "no/yes"}
    \label{fig:problem_illustration}
\end{figure}
Large Language Models (LLMs) and Vision-Language Models (VLMs) show impressive performance on benchmark question-answering tasks, even in some cases outperforming humans \cite{chowdhery2023palm}. However, upon closer inspection, their performance appears highly contingent on the input, and can drastically change with minor, superficial modifications. Recent studies have demonstrated this instability -- revealing model preferences for specific prompt orderings or biases towards particular answer positions in question-answering scenarios \cite{zhao2021calibrate, zheng2023large, pezeshkpour2023large} and unjustly showing preference or bias against specific demographic groups \cite{doi:10.1126/science.aal4230}. Even widely-used models like GPT-4 exhibit a bias to the first presented answers when used as an evaluator \cite{zheng2023judging, wang2023large}. This instability can be traced back to the inherent bias in training data -- which can lead models to learn superficial patterns and associations \cite{5995347}. For instance, \citet{zhao2021calibrate} discovered that GPT-3 is biased towards generating tokens common in its pretraining distribution. Figure \ref{fig:problem_illustration} display some examples demonstrating how bias can manifest in large models.

Eliminating this type of instability and bias is challenging. Existing methods either lack effectiveness or are characterized by high costs and inefficiency. \citet{zheng2023large} and \citet{pezeshkpour2023large} revealed that bias to a certain option position in Multiple Choice Question (MCQ) tasks persists after incorporating in-context examples in the prompt. Instead of mitigating the bias, in-context examples alter the favored position. Solving this issue solely by relying on more data, as in standard fine-tuning or in-context learning, is unlikely unless we can ensure bias-free training data, an impractical, if not impossible, task \cite{liang2022advances, khosla2012undoing}. Post-hoc model intervention during inference has been explored to sidestep data requirements. Calibration methods \cite{zheng2023large, pezeshkpour2023large} seek to debias output token probabilities to prevent disproportionate allocation of probability mass to specific tokens. However, this approach necessitates multiple inferences for the same sample, creating a computational load that grows exponentially with the number of presented options. Moreover, solely intervening in the output space limits our degree of freedom for intervention. A parallel line of research identifies directions in model's internal representations that correspond to some desirable trait (e.g., factual correctness) \cite{li2023inferencetime, burns2022discovering,liu2023context}, and steer activation values towards this direction. Unfortunately, these works require label supervision, making them prone to adopt data biases when applied to model debiasing.

As such, we ask: \ul{\textit{can we mitigate bias directly in the model representation space and do so without labeled data?}} We propose $\SYSNAME$, an \textit{unsupervised inference-time} intervention method designed for this purpose. Our approach capitalizes on the observation that bias often manifests as simple association rules, such as \texttt{"the first option is likely to be correct"}. Leveraging this insight, we construct a set of possible association rules and build a set of demonstrations from unlabeled samples that exemplify these rules. Subsequently, we identify the directions in the model representation space corresponding to these rules. During inference, we shift activations away from these identified directions.

Despite not using any label information, $\SYSNAME$ reduces three instruction-tuned model performance variability across different option ordering on three benchmark tasks: two yes/no questions and one large MCQ dataset by an average of 10.86\% accuracy points and 12.95 score points. Remarkably, it not only \textbf{outperforms a supervised baseline} with 100 labels but also matches the performance achieved with 500 labels. 
We systematically analyze $\SYSNAME$ to understand the bias directions it discovers. We empirically show that steering bias directions \textit{does not negatively} impact base model performance, and identified bias direction is generalizable across different datasets with the same task. Additionally, only a small number of unlabeled sample demonstrations are sufficient to identify these bias directions. 

To summarize, our contributions include,
\begin{itemize}[leftmargin=*]
    \item We propose $\SYSNAME$, an \ul{\textit{unsupervised}} inference-time activation steering algorithm to mitigate foundation model bias. 
    \item We demonstrate that $\SYSNAME$ can effectively address the instability concerning option ordering in question-answering tasks. Furthermore, our findings demonstrate that the bias direction pinpointed by $\SYSNAME$ is generalizable across datasets with the same task.
    \item Extensive experimental evidence shows improvement on three instruction-tuned models, with reduced performance variability by 10.86\% accuracy points across three datasets.

\end{itemize}

\section{Preliminaries}
\label{sec:preliminaries}
In this section, we describe our problem setup in more detail and briefly describe the transformer architecture \cite{vaswani2017attention} to set notation and context.
\subsection{Problem Statement} 
\label{sec:setup}
Given a model $T$ and a set of questions $q$ and $q^{\prime}$, each representing a slight variant of the same prompt with non-meaning altering changes, for instance:
\begin{align*}
& q = \text{Is Yosemite in California? Answer \underline{\textit{yes or no}}} \\ & q^{\prime} = \text{Is Yosemite in California? Answer \underline{\textit{no or yes}}}
\end{align*}
Our goal is to ensure consistent model outputs $T(q) = T(q')$ with the given model $T$. The question permutation set, denoted as $q, q', q'', ... $, can also include any number of variations to the same question. For example, multiple-choice questions (MCQs) with shuffled options, or, in multi-modal settings, an image-question pair with changes to the image that should not influence the model's answer. 
For simplicity, we use $q$ and $q'$ as our running example, though $\SYSNAME$ is applicable to any number of variants, as we will later show in Section $\ref{sec:experiments}$.

\subsection{Model Architecture} Our approach is applicable to any transformer-based \cite{vaswani2017attention} models. We follow the idea presented in \cite{elhage2021mathematical}, viewing the inner computations of transformers as a series of ``residual blocks". During inference, the token embedding layer initiates the ``residual stream" by projecting tokens into a high-dimensional space $x \in \mathbbm{R}^{DH}$. Subsequent layers then handle the flow of information within the stream—projecting into its own subspace, performing computations, and reprojecting its output back to the original space before feeding it to the next layer.

After the token embedding layer, each layer $l$ is composed of a multi-head-attention (MHA) module followed by a multi-layer perceptron (MLP). The MHA module consists of $H$ attention heads, with each head $h$ independently performing linear computations in parallel.

We leverage the findings from \cite{elhage2021mathematical} showing the equivalence of stacking $h$ outputs in the original Transformers paper, with taking the sum of $h$ outputs and projecting it back to the residual stream. For an input $x$, we can write the MHA computation as:
\begin{align*}
    MHA(x) = x + \sum _{h=1}^{H} W_{Ol}^{h}Att_l^h(W_{Vl}^hx),
\end{align*}
where $H$ is the set of attention heads $h$ in layer $l$, $W_{Ol}^h \in \mathbbm{R}^{DH \times D}$ is the output weight matrix, $W_{Vl}^h \in \mathbbm{R}^{D \times DH}$ the projection matrix to the attention head space, and $Att$ is the attention operator (which encapsulates multiplication to key and query weight matrices). To have as many linear properties as possible, we design our intervention after the $Att$ output and before $W_{Ol}^h$. We denote attention head activation value of any input $x$ at head $h$ of layer $l$ as $\theta_{h,l}^x \in \mathbbm{R}^D$.

\section{$\SYSNAME$: Unsupervised Inference-Time Debiasing}
\label{sec:method}
\begin{figure*}[ht!]
    \centering
    \includegraphics[width=\textwidth]{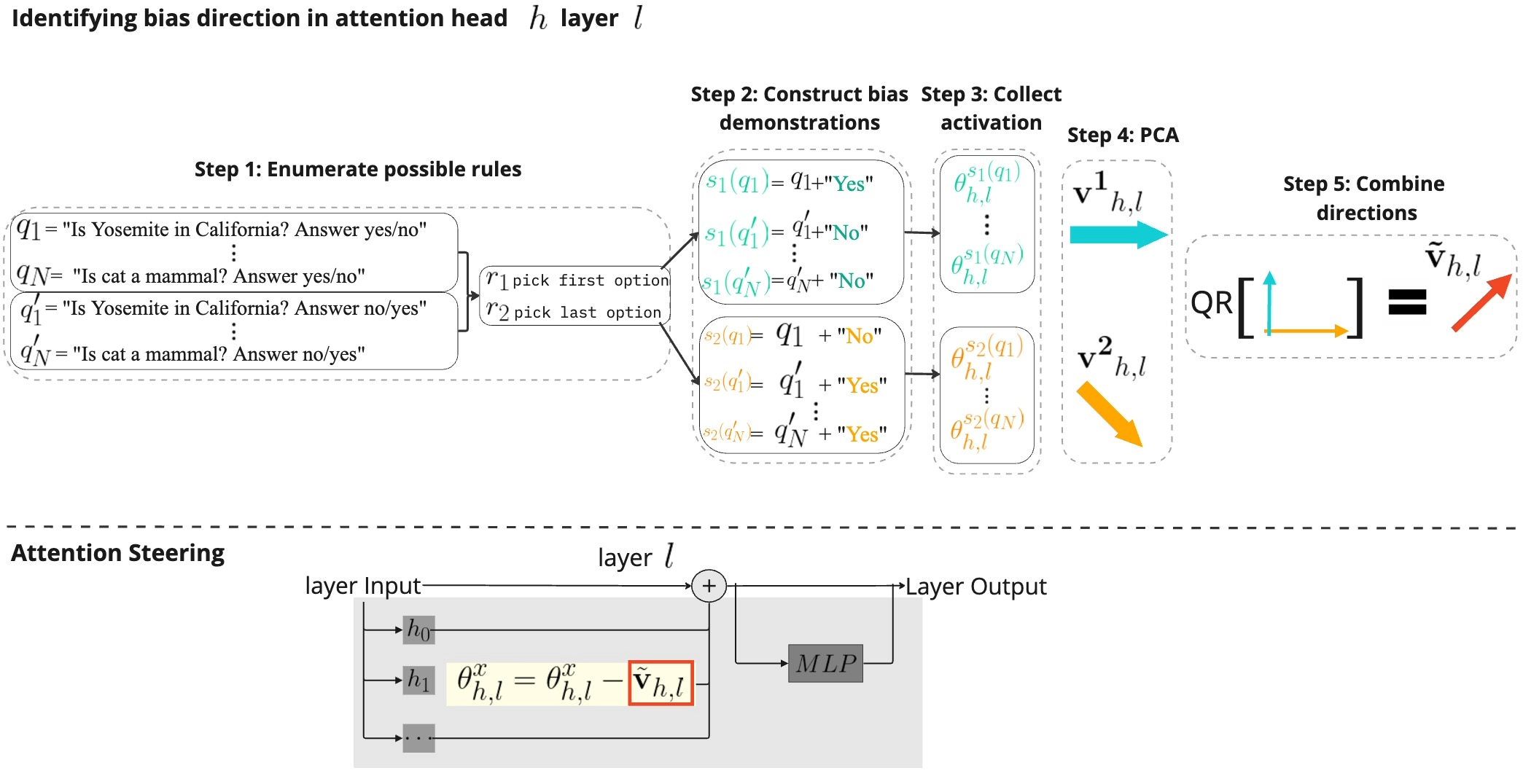}
    \caption{$\SYSNAME$ finds bias directions $\tilde{\textbf{v}}_{h,l}$ (top) and steer attention head values (bottom) away from it during inference.}
    \label{fig:main_diagram}
\end{figure*}
We are ready to describe $\SYSNAME$: an unsupervised inference-time debiasing method. 

In Section \ref{sec_intro}, we describe how models adopt unwanted bias towards superficial characteristics in the input, such as option location. Our goal is to reduce bias \ul{\textit{directly in the model's internal representations without using labeled data}}. To achieve this, we exploit the fact that bias often consists of spurious simple rules, like \texttt{always choose the first option}. Our method, $\SYSNAME$, leverages this fact by mimicking bias behaviors (e.g., answering both $q$ and $q'$ with the first option) and collecting their representations. Next, we identify directions in the model's representation space that encapsulate such bias and steer activation values away from it during inference. Our approach is illustrated in Figure \ref{fig:main_diagram} and summarized in Algorithm \ref{alg:fairfix}.

\subsection{Enumerating Bias Association Rules}  We capitalize on the intrinsic property of bias, that it comprises of simple, imitable rules. To leverage this property, we begin by enumerating the possible association rules the model might adopt. In question answering task with $m$ options, this is straightforward: $\left \{ \texttt{"Always choose the } j \texttt{th option"}\right \}, \forall j \in \left \{1,...m\right \}$. We refer to this set as the \textit{bias rule set} $\mathbbm{r} = \left \{ r_1,r_2,...,r_m\right \}$. In our running example, the set has two rules: the bias to the first option and to the last option, denoted as: $\mathbbm{r} = \left \{ r_1,r_2 \right \}$. 
Although models may adopt more intricate bias rules, like preferring some option tokens in specific positions, we assume these are highly correlated and influenced by position.

\subsection{Constructing Bias Demonstrations} For each rule, we then construct demonstrations by collecting \textit{a set of question-answer pairs that mimic the bias}. Recall that in our running example, $q$'s first option is ``yes" and last option is ``no'', and $q'$'s option order is flipped, this yields:
\begin{align*}
& s_1(q) = q + \text{``Answer: yes"} \quad s_2(q) = q + \text{``Answer: no"} \\
& s_1(q') = q' + \text{``Answer: no"} \quad s_2(q') = q' + \text{``Answer: yes"},
\end{align*}
where $s_j(q)$ is the demonstration of the $j$th rule with question $q$. We will refer to the set of demonstrations as the \textit{demonstration set}. In this specific example, we have two demonstration sets: $\mathcal{S}_1=\left \{ s_1(q), s_1(q') \right \}$ and $\mathcal{S}_2 =\left \{  s_2(q), s_2(q') \right \}$. 

Transitioning from our synthetic example, let's now form a bias rule set and demonstration set for a dataset of yes/no questions $\left \{ q_1, ..., q_N \right \}$. Similar to the synthetic case, our bias rule set is $\mathbbm{r} = \left \{ r_1,r_2 \right \}$, representing biases towards the first and last options. The resulting demonstration sets are: \begin{align*}
\mathcal{S}_1 = \left \{ s_1(q_i), s_1(q'_i)\right \} \\
\mathcal{S}_2 = \left \{ s_2(q_i), s_2(q'_i)\right \} \\
\forall i \in \left \{ 1, \dots N \right \}    
\end{align*}.
Construction of bias rule set and demonstration set is conducted without any label supervision. For simplicity, as an example we present the straightforward case of yes/no questions with two possible bias rules. Extending this to cases like multiple-choice questions (MCQ), where we may have more than two rules, is straightforward. We elaborate on MCQ rule and demonstration construction in Appendix \ref{appendix:mcq_demonstration} and present $\SYSNAME$ performance on both yes/no questions and MCQ in Section \ref{sec:experiments}.

\subsection{Identifying Bias Directions from Demonstrations} Our goal in this step is to identify directions in the attention head space for each bias rule. Given a bias rule set $\mathbbm{r}$ with $m$ rules, the input to this step is the demonstration sets $\left \{ \mathcal{S}_1, ..., \mathcal{S}_m \right \}$, and the output is $m$ bias directions per attention head per layer. 

Specifically, we collect the attention activation values of all demonstration set $\mathcal{S}^j$ elements at the last token position. This will result in a collection of latent state vectors denoted as
\begin{equation*}
    \textbf{H}_{h,l}^j := \left [ \theta_{h,l}^{s_j(q_1)} \middle| \ldots \middle| \theta_{h,l}^{s_j(q_N)}\right ]^T,  
\end{equation*}
with $\textbf{H}_{h,l}^j \in \mathbbm{R}^{N \times D}$. Next, we find the direction corresponding to each bias rule by performing PCA on $\textbf{H}_{h,l}^j$, and take the first principal direction ($\text{PCA1}$). We denote the resulting bias direction at attention head $h$ of layer $l$ as $\textbf{v}_{h,l}^j \in \mathbbm{R}^D$, which can be expressed as 
\begin{equation}
\textbf{v}_{h,l}^j = \text{PCA1}(\textbf{H}_{h,l}^j)
\end{equation}

\subsection{Combining Multiple Bias Directions}
\label{sec:method_combine_bias}
In each attention head $h$ of layer $l$, we now have a set of vectors $\left \{ \textbf{v}_{h,l}^1,..,\textbf{v}_{h,l}^m\right \}$, where each corresponds to the directions in activation space that best represents a bias rule in $\mathbbm{r}$. The remaining question is how to aggregate these directions so that steering them away from activation values during inference does not cause excessive disruption. Summation may produce large values with a large $m$, and a few highly correlated directions can dominate the average. To circumvent this, we use standard matrix decomposition methods, specifically QR decomposition, to obtain the orthonormal basis of the directions and take their average.
\begin{equation}
    \tilde{\textbf{v}}_{h,l} = \frac{1}{m} \text{QR}\left [\textbf{v}^1_{h,l}
                        \middle|\ldots \middle|\textbf{v}^m_{h,l}
                        \right ] ^T
\end{equation}
This way, we remove correlations between bias directions.
\subsection{Shifting Activation during Inference} During inference, we steer the activation values in the last token away from $\tilde{\textbf{v}}_{h,l}$, written as
\begin{equation}
      MHA(x) = x + \sum _{h=1}^{H} W_{Ol}^{h}(Att_l^h(W_{Vl}^hx) - \alpha \tilde{\textbf{v}}_{h,l}),
\end{equation}
where $\alpha$ is a hyperparameter that controls the strength of intervention. Finally, to preserve the model’s original capabilities as much as possible, we normalize the updated
latent states to match the $l_2$ norm of the latent states before the update.

\subsection{Selecting Attention Heads to Intervene} To be minimally invasive, we select the top $K$ attention heads with the highest average projected values in the first principal components. Intuitively, this is equivalent to selecting heads whose bias direction is mostly captured by the first principal component, and thus, our procedure will be the most effective. In section \ref{sec:no_tuning}, we empirically show that setting $\alpha = 1$ and setting $K = $ all heads still produces improvement over baselines, but we can get a noticeable boost by hyperparameter tuning.
\begin{algorithm}[H]
\small
        \caption{Identifying bias direction with \textsc{$\SYSNAME$}} \label{alg:fairfix}
        \begin{flushleft}
	\begin{algorithmic}[1]
		\STATE \textbf{Parameters:}
            Foundation model with $l$ layers and $h$ attention heads per layer, Dataset of questions with different prompt orderings $\left \{(q_1,...,q_N), (q'_1,...,q'_N), (q''_1,...,q''_N), ... \right \} $, strength hyperparameter $\alpha$  \\
            \STATE Enumerate set of rules $\mathbbm{r} = \left \{ r_1,r_2,..,r_m \right \}$
            
            \FOR{$j \in \left \{1,2,..m \right \}$}
            \STATE Construct demonstration sets $\mathcal{S}_j = \left \{ s_j(q_i)\right \} \forall i \in \left \{1,\ldots,N\right \}$
            \STATE Collect attention head values $\textbf{H}_{h,l}^j$
            \STATE Identify direction  $ \textbf{v}_{h,l}^j = \text{PCA1}(\textbf{H}_{h,l}^j)$
            \ENDFOR         
            \STATE Combine directions $\tilde{\textbf{v}}_{h,l} = \frac{1}{m} \text{QR}\left [
                        \textbf{v}^1_{h,l}\middle|\ldots \middle|
                        \textbf{v}^m_{h,l}
                        \right ]$
            \STATE Attention steering $MHA(x) = x + \sum _{h=1}^{H} W_{Ol}^{h}(Att_l^h(W_{Vl}^hx) - \alpha \tilde{\textbf{v}}_{h,l})$
            \STATE \textbf{Returns:} (intervened) foundation model
	\end{algorithmic} 
        \end{flushleft}
   
\end{algorithm}


\section{Empirical Evaluation}
\label{sec:experiments}
\begin{table*}[t!]
    \centering
   \small
   \setlength\tabcolsep{3.5pt}
    \begin{tabular}{llcccccccc}
        \toprule
        \multirow{2}{*}{Dataset} & \multirow{2}{*}{Model} & \multicolumn{2}{c}{Vanilla} & \multicolumn{2}{c}{ITI (supervised 100)} & \multicolumn{2}{c}{ITI (supervised 500)} & \multicolumn{2}{c}{\textbf{$\SYSNAME$} (unsupervised)} \\ 
        \cmidrule(lr){3-4} \cmidrule(lr){5-6} \cmidrule(lr){7-8}\cmidrule(lr){9-10} 
        && Avg\%($\uparrow$) & Std($\downarrow$) & Avg\%($\uparrow$) & Std($\downarrow$) & Avg\%($\uparrow$) &Std($\downarrow$)& Avg\%($\uparrow$) &Std($\downarrow$) \\
        \toprule
        \multirow{3}{*}{ScienceQA} & LLaVA (13B) & \underline{64.28\%} & 0.024 & 64.22\% & 0.029 & 64.05\% & \cellcolor{green!20}{\textbf{0.015}} & \cellcolor{blue!20}{\textbf{65.46\%}} & \underline{0.017} \\
        & IDEFICS (9B) &  \cellcolor{blue!20}{\textbf{58.99}}\% & 0.181 & 54.74\% & \cellcolor{green!20}{\textbf{0.079}} & 56.03\% & \underline{0.125}  & \underline{58.70\%} & 0.152 \\
        & InstructBLIP (13B) & \underline{56.32\%} & \underline{0.213} & 55.43\% & 0.257 & 55.38\% & 0.259 & \cellcolor{blue!20}{\textbf{56.92\%}} & \cellcolor{green!20}{\textbf{0.092}} \\
        \midrule
        \multirow{3}{*}{MME} & LLaVA (13B) & 1333.43 & 65.42 & \cellcolor{blue!20}{\textbf{1350.19}} & 47.42 & 1334.87 & \underline{10.43} & \underline{1333.56} & \cellcolor{green!20}{\textbf{2.72}} \\
        & IDEFICS (9B) & \cellcolor{blue!20}{\textbf{1044.70}} & 49.92 & 1011.31 & 6.00 & 1023.55 & \cellcolor{green!20}{\textbf{1.78}} & \underline{1035.83} & \underline{2.39} \\
        & InstructBLIP (13B) & 1175.85 & 15.09 & \underline{1184.30} & \underline{5.58} & 1180.37 & \cellcolor{green!20}{\textbf{0.75}} &  \cellcolor{blue!20}{\textbf{1185.93}} & 15.03 \\
        \midrule
        \multirow{3}{*}{VGR} & LLaVA (13B) & 71.04\%\ & 0.126 & 65.91\% & \underline{0.079} & \cellcolor{blue!20}{\textbf{71.63\%}} & 0.091 & \underline{71.46\%} & \cellcolor{green!20}{\textbf{0.054}} \\
        & IDEFICS (9B) & \cellcolor{blue!20}{\textbf{52.59\%}} & 0.151 & 52.17\% & 0.286 & 50.53\% & \cellcolor{green!20}{\textbf{0.051}} & \underline{52.07\%} & \underline{0.060} \\
        & InstructBLIP (13B) & \cellcolor{blue!20}{\textbf{51.38\%}} & 0.242 & 50.32\% & 0.303 & 50.27\% & \underline{0.172}  & \underline{50.31\%} & \cellcolor{green!20}{\textbf{0.006}}\\
        \bottomrule
    \end{tabular}
    \caption{Order bias results. Best method in \textbf{bold}, runner-up \underline{underlined}. We compare against ITI \cite{li2023inferencetime} with 100 and 500 labels}
    \label{tab:order_bias_exp}
\end{table*}


\begin{figure*}
\begin{subfigure}{0.35\textwidth}
  \includegraphics[width=\linewidth]{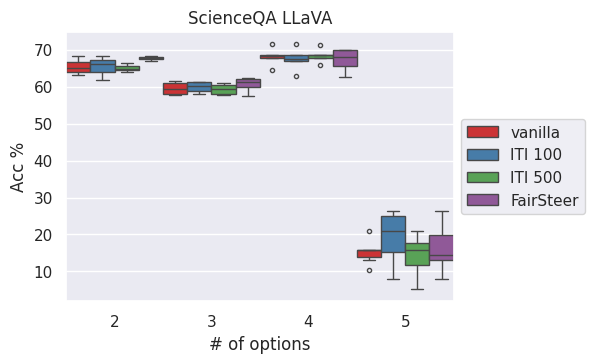}
\end{subfigure}
\begin{subfigure}{0.32\textwidth}
  \includegraphics[width=\linewidth]{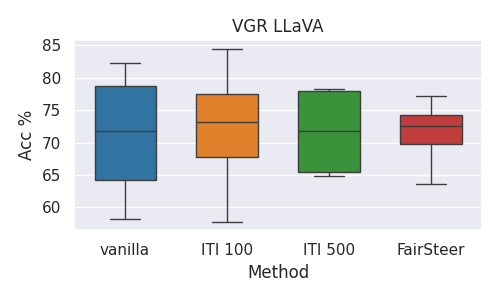}
\end{subfigure}
\begin{subfigure}{0.32\textwidth}
  \includegraphics[width=\linewidth]{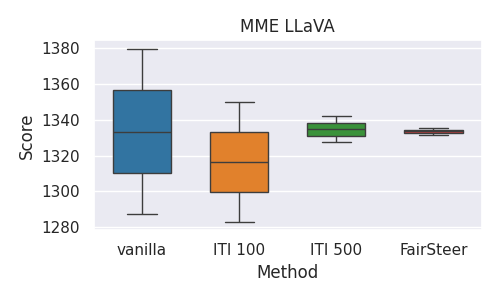}
\end{subfigure}
\caption{Left to right: ScienceQA, VGR, MME. $\SYSNAME$ reduces standard deviation across prompt ordering while maintaining average accuracy.}
\label{fig:main_results}
\end{figure*}

In this section, we empirically verify the effectiveness of $\SYSNAME$ with regard to:
\begin{itemize}[leftmargin=*]
    \item \textbf{Mitigating Order Bias} (Section \ref{sec:order_bias}):  Our unsupervised method mitigates model's tendency to choose options at specific positions and is \ul{\textit{competitive to (and sometimes outperforms) supervised methods} }\cite{li2023inferencetime}.
    \item \textbf{Generalization Capability} (Section \ref{sec:generalization}): We show that the bias direction identified by $\SYSNAME$ is generalizable across datasets with the same task.
\end{itemize}

\paragraph{Baselines.} We compare $\SYSNAME$ against vanilla inference of instruction-tuned models: LLaVA (13B) \cite{liu2023llava, liu2023improvedllava}, IDEFICS (9B) \cite{laurenccon2023obelisc} (open-source Flamingo \cite{alayrac2022flamingo}), and InstructBLIP (13B) \cite{dai2305instructblip}. For the order bias task, we include a comparison with Inference-Time Intervention (ITI) \cite{li2023inferencetime}, a supervised attention steering method.

\paragraph{Experimental Setup.} We use a separated unlabeled training set for finding bias directions and testing. Unless stated otherwise, we follow the default split for each dataset. The results in this section are based on 1000 random unlabeled samples from the separated training set, with the exception of the MME Benchmark \cite{fu2023mme}, where we use 100 samples due to its smaller size. We provide the full dataset and prompt details in Appendix \ref{appendix:datasets} and \ref{appendix:prompt}.

\subsection{Mitigating Order Bias}
\label{sec:order_bias}

\paragraph{Setup.} We test our method on three Multiple Choice Question (MCQ) and yes/no question-answering datasets: ScienceQA \cite{lu2022learn}, {MME Benchmark} \cite{fu2023mme}, and {Visual Genome Relation} (VGR) \cite{cocodataset}. On yes/no questions, we record the performance across different option orderings (i.e., ``answer with a yes/no'' and ``answer with a no/yes''). To test this bias in MCQ dataset, we employ the \textit{answer-moving attack} evaluation from \cite{zheng2023large}, by always moving the golden answers to a specific position. 

\paragraph{Metrics.} We report the average accuracy across option orders (Avg\%) and the corresponding standard deviation (Std). In the case of MME Benchmark, we adhere to the evaluation score proposed in the original work. Note that the score is not in percentage form, resulting in a higher scale of standard deviation. A model with less bias to option order will have a high Avg\% and low Std.

\paragraph{Results.} Table \ref{tab:order_bias_exp} shows that \textbf{$\SYSNAME$ significantly reduces bias to option order while often improving the average accuracy}. Remarkably, our unsupervised method surpasses ITI (supervised) performance on 100 labels and is comparable, and sometimes surpasses with 500 labels. This suggests that our method effectively reduces bias influence on the model's predictions, forcing the model to rely on relevant knowledge.

Figure \ref{fig:main_results} shows the non-averaged results on the LLaVA model. On the VGR and MME datasets, $\SYSNAME$ significantly reduces the standard deviation (as indicated by the box plot height) while maintaining the average accuracy. A similar pattern is observed on the ScienceQA dataset, particularly for options 2 and 3. However, $\SYSNAME$'s impact is more limited for a larger number of options. We hypothesize that this is caused by $\SYSNAME$'s simple averaging approach to combine multiple bias directions (Section \ref{sec:method_combine_bias}), as the number of bias directions grows with the number of options. This indicates potential areas for improvement in this part of our approach. Comprehensive non-averaged results for all models are provided in Appendix \ref{appendix:supp_results}.

\begin{table*}[ht!]
    \centering
    \begin{tabular}{llccccccccc}
        \toprule
        \multirow{2}{*}{Model} & \multirow{2}{*}{Dataset} & \multicolumn{2}{c}{Vanilla} & \multicolumn{2}{c}{Original Direction (OD)} & \multicolumn{2}{c}{Transferred Direction (TD)} \\ 
        \cmidrule(lr){3-4}  \cmidrule(lr){5-6}  \cmidrule(lr){7-8} 
        & & Avg\%($\uparrow$) & Std($\downarrow$) & Avg\%($\uparrow$) & Std($\downarrow$) & Avg\%($\uparrow$) & Std($\downarrow$)  \\
        \toprule
        \multirow{2}{*}{LLaVA (13B)} & VGR & 71.04\% & 0.126 & \underline{71.46\%} & \underline{0.054} & \cellcolor{blue!20}{\textbf{73.19\%}} & \cellcolor{green!20}{\textbf{0.015}} \\
        & MME & \underline{1333.43} & 65.42 & \cellcolor{blue!20}{\textbf{1333.56}} & \cellcolor{green!20}{\textbf{2.72}} & 1305.57 & \underline{56.54} \\
        \midrule
        \multirow{2}{*}{IDEFICS (9B)} & VGR & \underline{52.59\%} & 0.151 & 52.07\% & \cellcolor{green!20}{\textbf{0.060}} & \cellcolor{blue!20}{\textbf{52.76\%}} & \underline{0.144}  \\
        & MME & 1044.70 & \underline{49.92} & 1035.83 & \cellcolor{green!20}{\textbf{2.39}} & \cellcolor{blue!20}{\textbf{1060.94}} & \underline{11.35} \\
        \midrule
        \multirow{2}{*}{InstructBLIP (13B)} & VGR & \cellcolor{blue!20}{\textbf{51.38\%}} & 0.242 & \underline{50.31\%} & \cellcolor{green!20}{\textbf{0.006}} & 50.10\% & \underline{0.212} \\
        & MME & \underline{1175.85} & \underline{15.09} & \cellcolor{blue!20}{\textbf{1185.93}} & \cellcolor{green!20}{\textbf{15.03}} & 844.37 & 54.78 \\
        \bottomrule
    \end{tabular}
    \caption{$\SYSNAME$ generalization performance. $\SYSNAME$ with original direction (OD) uses direction identified using the dataset, $\SYSNAME$ with transferred direction (TD) uses direction identified from another dataset of the same task. Best \textbf{bolded}, second best \underline{underlined}.}
    \label{tab:generalization}
    \vspace{-1em}
\end{table*}
\subsection{Generalization Capability}
\label{sec:generalization}
\paragraph{Setup.} Next, we study the generalization property of bias directions identified by $\SYSNAME$. Specifically, we test whether attention steering using directions identified from a different dataset with the same task (e.g., VGR and MME) can produce similar results as using identified direction from the same dataset. Intuitively, the order bias problem is not domain-specific, so the bias directions should generalize.

\paragraph{Results.} The results in Table \ref{tab:generalization} illustrate that $\SYSNAME$ with directions transferred from another dataset (TD) consistently enhances the base model, demonstrating reduced performance variance and improved average accuracy. However, exceptions exist in cases where $\SYSNAME$ with directions from the same dataset (OD) fails to produce significant improvement, as observed, for instance, in the case of InstructBLIP on the MME dataset. This shows that \textbf{the bias direction identified by $\SYSNAME
$ generalizes} across datasets with the same task.

\section{Analysis}
\label{sec:analysis}
This section presents analyses of $\SYSNAME$ components and provides empirical characterizations.
\begin{figure}[h]
    \centering
    \includegraphics[width=.5\textwidth]{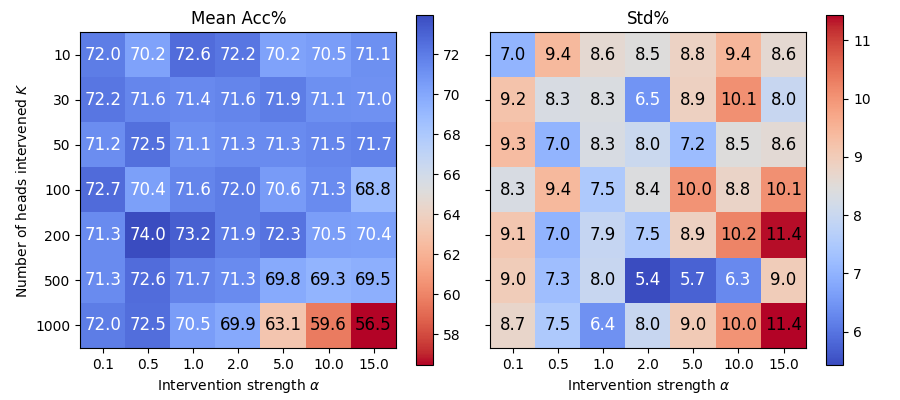}
    \caption{Effect of hyperparameters $\alpha$ (x-axis) and number of intervened attention heads $K$ (y-axis). Left: Acc\%; Right: Std\%. Performance recorded for VGR dataset.
    }
    \label{fig:hparams_ablation}
    \vspace{-1em}
\end{figure}
\subsection{Sensitivity to hyperparameters $\alpha$ and $K$} In Figure \ref{fig:hparams_ablation} (left), we observe that Avg Acc\% is insensitive to hyperparameters, shown by similar accuracies across different parameter values. Once both $\alpha$ and $K$ are large ($\alpha >= 5$ $K >= 1000)$, the steering changes the original activation values too much, causing accuracy to decline.

The standard deviation plot (Figure \ref{fig:hparams_ablation} (right)) shows good performance regime below the diagonal for $\alpha>$ 0.1. This shows that a combination of more number of heads intervened, with moderate $\alpha$ values yields the most effective bias reduction. Notably, across various parameter settings, the Std\% consistently matches or surpasses ITI's and outperforms the vanilla model.

\subsection{Can we get away without hyperparameters?}
\label{sec:no_tuning}
\begin{table*}
    \centering
    \resizebox{.75\textwidth}{!}{
    \begin{tabular}{llcc|cc|ccc}
        \toprule
        \multirow{2}{*}{Dataset} &
        \multirow{2}{*}{Model} & \multicolumn{2}{c}{Vanilla}& \multicolumn{2}{c}{With tuning}& \multicolumn{2}{c}{No tuning} \\ 
        \cmidrule(lr){3-4}  \cmidrule(lr){5-6} \cmidrule(lr){7-8}
        & & Avg\%($\uparrow$) & Std($\downarrow$) & Avg\%($\uparrow$)  & Std($\downarrow$) & Avg\%($\uparrow$)  & Std($\downarrow$) \\
        \toprule
        \multirow{3}{*}{SQA} & LLaVA & \underline{64.28\%} & 0.024 & \textbf{65.46\%} & \textbf{0.017} & 62.00\% & \textbf{0.017}  \\
        & IDEFICS & \underline{56.32\%} & 0.213 & \textbf{56.92\%} & \textbf{0.092} & 53.54\% & \underline{0.210}  \\
        & InstructBLIP & 56.32\%& 0.213 & \underline{56.92\%} & \textbf{0.092} & \textbf{57.75\%} & \underline{0.148} \\
        \midrule 
        \multirow{3}{*}{VGR} & LLaVA& \underline{71.04\%} & 0.126 & \textbf{71.46\%} & \underline{0.054} & 67.03\% & \textbf{0.023} \\
         & IDEFICS & \textbf{52.59\%} & 0.151 & \underline{52.07\%} & \underline{0.060} & 51.16\% & \textbf{0.023} \\
         & InstructBLIP & \underline{51.38\%} & 0.242 & 50.31\% & \textbf{0.006} & \textbf{52.06\%} & \underline{0.203}  \\
         \bottomrule
    \end{tabular}}
    \caption{Comparison with $\SYSNAME$ without tuning $\alpha$ and $K$. Best numbers in \textbf{bold}, second best \underline{underlined} $\SYSNAME$ without tuning still produces improvement over the vanilla model.
    }
    \label{tab:no_tuning}
\end{table*}

Here we test $\SYSNAME$ performance without any hyperparameter tuning. We intervene all attention heads ($K = \text{all}$), and set the strength hyperparameter $\alpha = 1$. Table \ref{tab:no_tuning} shows that hyperparameter tuning produces the best results. However, it is worth noting that $\SYSNAME$ without tuning still improves the performance of the vanilla model.

\subsection{How do bias directions in the activation space look?}
\begin{figure}[ht!]
    \centering
    \includegraphics[width=.5\textwidth]{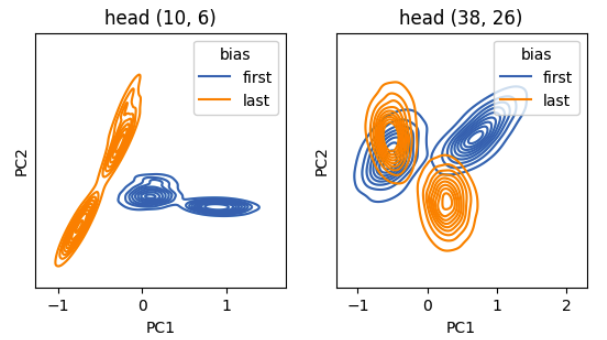}
    \caption{Kernel density estimate plots of $\SYSNAME$-identified bias directions on the VGR dataset, projected onto the first 2 PCs.
    }
    \label{fig:bias_viz}
    \vspace{-0.5em}
\end{figure}
Figure \ref{fig:bias_viz} illustrates the geometry of bias directions identified by $\SYSNAME$ before the combination step, projected onto the first two principal components. Two crucial observations emerge. Firstly, there is minimal overlap between distinct bias directions, indicating their simplicity and separability with just two principal components. Secondly, the maximum variance directions in the two density plots (blue and orange) are nearly orthogonal. This underscores the advantage of decomposing `bias' into multiple rules rather than treating it as a singular component.

\subsection{How many unlabeled samples do we need?}
\begin{figure}[h]
    \centering
    \includegraphics[width=.45\textwidth]{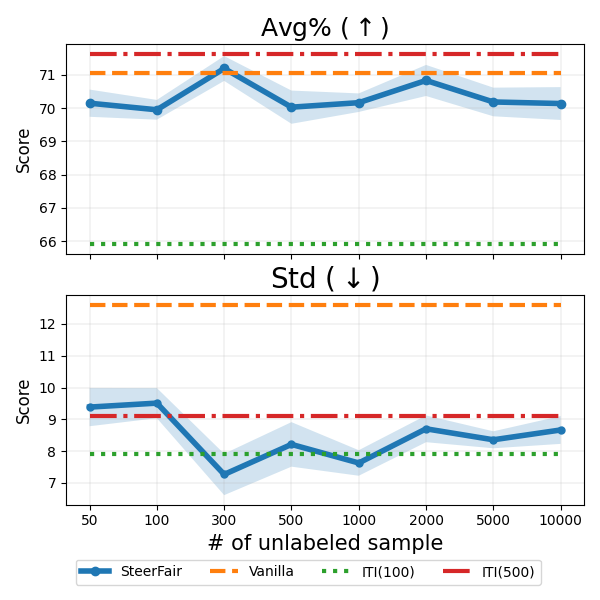}
    \caption{Impact of number of unlabeled sample $N$ (x-axis) on performance (y-axis). Performance across 10 random seeds.}
    \label{fig:samlpe_ablation}
\end{figure}
\begin{figure}
    \centering
    \includegraphics[width=.45\textwidth]{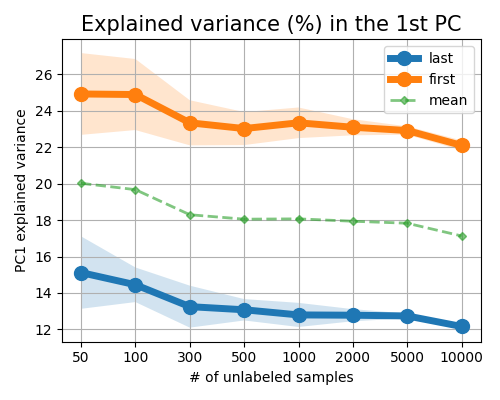}
    \caption{Explained variance ratio in the first principal component (PC) of bias to the first (blue) and last options (orange) directions identified by $\SYSNAME$ (y-axis) across the number of unlabeled samples $N$ (x-axis). Performance across 10 random seeds.
    }
    \label{fig:pca_explanation}
    \vspace{-1em}
\end{figure}
We vary the number of unlabeled samples $N$ to find bias direction and see the impact on performance in Figure \ref{fig:samlpe_ablation}. On the top figure, it is evident that Avg\% remains relatively stable, exhibiting performance fluctuations within the range of $\pm 2\%$ across different values of $N$. Intriguingly, even with a small sample size ($N < 100$), the accuracy is preserved, underscoring the non-intrusive nature of our intervention technique. On the bottom figure, we observe an interesting trend: there is no (negative) correlation between $N$ and resulting Std. The lowest standard deviation is achieved between $N=300$ and $N=1000$. This suggests that bias direction can be effectively approximated with a relatively small number of key samples.

It is surprising that larger $N$ does not yield better performance.  We hypothesize this is because more samples include more information, and thus bias direction extracted is more noisy. To validate this hypothesis, we examine the captured variance ratio in the first principal component (PC) across different $N$ values in Figure \ref{fig:pca_explanation}. It is evident that for both small and large values of $N$ ($N < 100$ and $N > 1000$), the captured variance in the first PC, and hence the informativeness of $\SYSNAME$ bias direction, is smaller than for $300 \leq N \leq 1000$, supporting our hypothesis.

\section{Related Work}
\label{sec:related_work}
\subsection{Foundation Model Bias and Robustness}
Foundation model robustness is a heavily studied area \cite{adila2023zero, zhang2022contrastive, yang2023mitigating}. The majority of these works are tailored to embedding-based models, such as CLIP, and are designed for scenarios with access to single embeddings per sample. This poses challenges when extending these approaches to next-word prediction transformer-based models, where the notion of single embedding is less straightforward. 

On next-word prediction models, works such as \cite{zheng2023large, pezeshkpour2023large} address vulnerabilities to superficial prompt modifications \cite{zhao2021calibrate, zheng2023judging}, such as non-meaning altering changes in token order. Specifically targeting the model's inclination to favor particular options in MCQ settings, \cite{zheng2023large} and \cite{pezeshkpour2023large} propose methods involving the calibration of model probabilities at the option token level, with final predictions based on the calibrated output. Our approach, $\SYSNAME$, distinguishes itself in two crucial aspects: (1) it allows more flexible intervention by modifying the model's internal activations, and (2) $\SYSNAME$ is versatile, applicable to a broader types of tasks and biases.
\subsection{Extracting LLM Knowledge in the Latent Space}
Progress has been made in extracting and understanding latent knowledge of LLMs. \citet{burns2022discovering} find truthfulness direction by identifying a direction in the model's internal representation such that the probabilities assigned to truthful vs. non-truthful answers adhere to logical consistencies. \citet{gurnee2305finding} study how
high-level interpretable features are represented in the models' internal workings. \citet{moschella2022relative} discover that learned representations remain invariant across stochastic factors in different training runs. \citet{li2022emergent} present evidence indicating that a GPT variant trained to generate moves in the Othello game learns a representation of game states. Furthermore, \citet{liu2023context} identify a direction in the latent space that effectively summarizes knowledge from in-context samples. Building upon insights from these techniques, we search for where and how bias is learned in the model's internal representation space.
\subsection{Modifying LLM Attention} 
Modifying LLM attention space has shown remarkable success in steering model behavior toward desirable traits without fine-tuning. This includes style transfer \cite{subramani2022extracting}, enhancing truthfulness \cite{li2023inferencetime}, achieving more controllable and effective in-context learning and transfer learning, as demonstrated in \cite{liu2023context} and \cite{shi2023refocusing}, and improving instruction-following capabilities \cite{zhang2023tell}. Ours shares a similar spirit with these works; we design an attention-steering mechanism specifically to mitigate model bias. Notably, our method leverages the inherent properties of the targeted desirable trait (i.e., reduced bias) rather than relying on label supervision.

\section{Conclusion}
\label{sec:conclusion}
We introduce $\SYSNAME$, an approach for unsupervised mitigation of bias in question-answering models, operating \textit{directly in the model representation space}. Our technique improves instruction-tuned model performance across various benchmark tasks, surpassing supervised baselines, and matching more amount labeled data scenarios. 
The comprehensive experimental analysis highlights the method's versatility, showing generalization properties of the identified bias directions and demonstrating efficacy with only a small number of unlabeled samples. The limitations of our work are discussed in Appendix \ref{appendix:limitations}.

\section*{Impact Statement}
This paper presents work whose goal is to advance the field of Machine Learning. There are many potential societal consequences of our work, none which we feel must be specifically highlighted here. As of now, we are not aware of additional potential societal impacts beyond the typical implications associated with LLMs in question-answering scenarios.



\bibliography{example_paper}

\begin{thebibliography}{33}
\providecommand{\natexlab}[1]{#1}
\providecommand{\url}[1]{\texttt{#1}}
\expandafter\ifx\csname urlstyle\endcsname\relax
  \providecommand{\doi}[1]{doi: #1}\else
  \providecommand{\doi}{doi: \begingroup \urlstyle{rm}\Url}\fi

\bibitem[Adila et~al.(2023)Adila, Shin, Cai, and Sala]{adila2023zero}
Adila, D., Shin, C., Cai, L., and Sala, F.
\newblock Zero-shot robustification of zero-shot models with foundation models.
\newblock \emph{arXiv preprint arXiv:2309.04344}, 2023.

\bibitem[Alayrac et~al.(2022)Alayrac, Donahue, Luc, Miech, Barr, Hasson, Lenc, Mensch, Millican, Reynolds, et~al.]{alayrac2022flamingo}
Alayrac, J.-B., Donahue, J., Luc, P., Miech, A., Barr, I., Hasson, Y., Lenc, K., Mensch, A., Millican, K., Reynolds, M., et~al.
\newblock Flamingo: a visual language model for few-shot learning.
\newblock \emph{Advances in Neural Information Processing Systems}, 35:\penalty0 23716--23736, 2022.

\bibitem[Burns et~al.(2022)Burns, Ye, Klein, and Steinhardt]{burns2022discovering}
Burns, C., Ye, H., Klein, D., and Steinhardt, J.
\newblock Discovering latent knowledge in language models without supervision.
\newblock \emph{arXiv preprint arXiv:2212.03827}, 2022.

\bibitem[Caliskan et~al.(2017)Caliskan, Bryson, and Narayanan]{doi:10.1126/science.aal4230}
Caliskan, A., Bryson, J.~J., and Narayanan, A.
\newblock Semantics derived automatically from language corpora contain human-like biases.
\newblock \emph{Science}, 356\penalty0 (6334):\penalty0 183--186, 2017.
\newblock \doi{10.1126/science.aal4230}.
\newblock URL \url{https://www.science.org/doi/abs/10.1126/science.aal4230}.

\bibitem[Chowdhery et~al.(2023)Chowdhery, Narang, Devlin, Bosma, Mishra, Roberts, Barham, Chung, Sutton, Gehrmann, et~al.]{chowdhery2023palm}
Chowdhery, A., Narang, S., Devlin, J., Bosma, M., Mishra, G., Roberts, A., Barham, P., Chung, H.~W., Sutton, C., Gehrmann, S., et~al.
\newblock Palm: Scaling language modeling with pathways.
\newblock \emph{Journal of Machine Learning Research}, 24\penalty0 (240):\penalty0 1--113, 2023.

\bibitem[Dai et~al.()Dai, Li, Li, Tiong, Zhao, Wang, Li, Fung, and Hoi]{dai2305instructblip}
Dai, W., Li, J., Li, D., Tiong, A., Zhao, J., Wang, W., Li, B., Fung, P., and Hoi, S.
\newblock Instructblip: Towards general-purpose vision-language models with instruction tuning. arxiv 2023.
\newblock \emph{arXiv preprint arXiv:2305.06500}.

\bibitem[Elhage et~al.(2021)Elhage, Nanda, Olsson, Henighan, Joseph, Mann, Askell, Bai, Chen, Conerly, DasSarma, Drain, Ganguli, Hatfield-Dodds, Hernandez, Jones, Kernion, Lovitt, Ndousse, Amodei, Brown, Clark, Kaplan, McCandlish, and Olah]{elhage2021mathematical}
Elhage, N., Nanda, N., Olsson, C., Henighan, T., Joseph, N., Mann, B., Askell, A., Bai, Y., Chen, A., Conerly, T., DasSarma, N., Drain, D., Ganguli, D., Hatfield-Dodds, Z., Hernandez, D., Jones, A., Kernion, J., Lovitt, L., Ndousse, K., Amodei, D., Brown, T., Clark, J., Kaplan, J., McCandlish, S., and Olah, C.
\newblock A mathematical framework for transformer circuits.
\newblock \emph{Transformer Circuits Thread}, 2021.
\newblock https://transformer-circuits.pub/2021/framework/index.html.

\bibitem[Fu et~al.(2023)Fu, Chen, Shen, Qin, Zhang, Lin, Yang, Zheng, Li, Sun, et~al.]{fu2023mme}
Fu, C., Chen, P., Shen, Y., Qin, Y., Zhang, M., Lin, X., Yang, J., Zheng, X., Li, K., Sun, X., et~al.
\newblock Mme: A comprehensive evaluation benchmark for multimodal large language models.
\newblock \emph{arXiv preprint arXiv:2306.13394}, 2023.

\bibitem[Gurnee et~al.()Gurnee, Nanda, Pauly, et~al.]{gurnee2305finding}
Gurnee, W., Nanda, N., Pauly, M., et~al.
\newblock Finding neurons in a haystack: Case studies with sparse probing, may 2023.
\newblock \emph{URL http://arxiv. org/abs/2305.01610.→ p}, 9.

\bibitem[Khosla et~al.(2012)Khosla, Zhou, Malisiewicz, Efros, and Torralba]{khosla2012undoing}
Khosla, A., Zhou, T., Malisiewicz, T., Efros, A.~A., and Torralba, A.
\newblock Undoing the damage of dataset bias.
\newblock In \emph{Computer Vision--ECCV 2012: 12th European Conference on Computer Vision, Florence, Italy, October 7-13, 2012, Proceedings, Part I 12}, pp.\  158--171. Springer, 2012.

\bibitem[Lauren{\c{c}}on et~al.(2023)Lauren{\c{c}}on, Saulnier, Tronchon, Bekman, Singh, Lozhkov, Wang, Karamcheti, Rush, Kiela, et~al.]{laurenccon2023obelisc}
Lauren{\c{c}}on, H., Saulnier, L., Tronchon, L., Bekman, S., Singh, A., Lozhkov, A., Wang, T., Karamcheti, S., Rush, A.~M., Kiela, D., et~al.
\newblock Obelisc: An open web-scale filtered dataset of interleaved image-text documents.
\newblock \emph{arXiv preprint arXiv:2306.16527}, 2023.

\bibitem[Li et~al.(2022)Li, Hopkins, Bau, Vi{\'e}gas, Pfister, and Wattenberg]{li2022emergent}
Li, K., Hopkins, A.~K., Bau, D., Vi{\'e}gas, F., Pfister, H., and Wattenberg, M.
\newblock Emergent world representations: Exploring a sequence model trained on a synthetic task.
\newblock \emph{arXiv preprint arXiv:2210.13382}, 2022.

\bibitem[Li et~al.(2023)Li, Patel, Viégas, Pfister, and Wattenberg]{li2023inferencetime}
Li, K., Patel, O., Viégas, F., Pfister, H., and Wattenberg, M.
\newblock Inference-time intervention: Eliciting truthful answers from a language model, 2023.

\bibitem[Liang et~al.(2022)Liang, Tadesse, Ho, Fei-Fei, Zaharia, Zhang, and Zou]{liang2022advances}
Liang, W., Tadesse, G.~A., Ho, D., Fei-Fei, L., Zaharia, M., Zhang, C., and Zou, J.
\newblock Advances, challenges and opportunities in creating data for trustworthy ai.
\newblock \emph{Nature Machine Intelligence}, 4\penalty0 (8):\penalty0 669--677, 2022.

\bibitem[Lin et~al.(2014)Lin, Maire, Belongie, Bourdev, Girshick, Hays, Perona, Ramanan, Doll{'{a} }r, and Zitnick]{cocodataset}
Lin, T., Maire, M., Belongie, S.~J., Bourdev, L.~D., Girshick, R.~B., Hays, J., Perona, P., Ramanan, D., Doll{'{a} }r, P., and Zitnick, C.~L.
\newblock Microsoft {COCO:} common objects in context.
\newblock \emph{CoRR}, abs/1405.0312, 2014.
\newblock URL \url{http://arxiv.org/abs/1405.0312}.

\bibitem[Liu et~al.(2023{\natexlab{a}})Liu, Li, Li, and Lee]{liu2023improvedllava}
Liu, H., Li, C., Li, Y., and Lee, Y.~J.
\newblock Improved baselines with visual instruction tuning, 2023{\natexlab{a}}.

\bibitem[Liu et~al.(2023{\natexlab{b}})Liu, Li, Wu, and Lee]{liu2023llava}
Liu, H., Li, C., Wu, Q., and Lee, Y.~J.
\newblock Visual instruction tuning, 2023{\natexlab{b}}.

\bibitem[Liu et~al.(2023{\natexlab{c}})Liu, Xing, and Zou]{liu2023context}
Liu, S., Xing, L., and Zou, J.
\newblock In-context vectors: Making in context learning more effective and controllable through latent space steering.
\newblock \emph{arXiv preprint arXiv:2311.06668}, 2023{\natexlab{c}}.

\bibitem[Lu et~al.(2022)Lu, Mishra, Xia, Qiu, Chang, Zhu, Tafjord, Clark, and Kalyan]{lu2022learn}
Lu, P., Mishra, S., Xia, T., Qiu, L., Chang, K.-W., Zhu, S.-C., Tafjord, O., Clark, P., and Kalyan, A.
\newblock Learn to explain: Multimodal reasoning via thought chains for science question answering.
\newblock In \emph{The 36th Conference on Neural Information Processing Systems (NeurIPS)}, 2022.

\bibitem[Moschella et~al.(2022)Moschella, Maiorca, Fumero, Norelli, Locatello, and Rodola]{moschella2022relative}
Moschella, L., Maiorca, V., Fumero, M., Norelli, A., Locatello, F., and Rodola, E.
\newblock Relative representations enable zero-shot latent space communication.
\newblock \emph{arXiv preprint arXiv:2209.15430}, 2022.

\bibitem[Pezeshkpour \& Hruschka(2023)Pezeshkpour and Hruschka]{pezeshkpour2023large}
Pezeshkpour, P. and Hruschka, E.
\newblock Large language models sensitivity to the order of options in multiple-choice questions.
\newblock \emph{arXiv preprint arXiv:2308.11483}, 2023.

\bibitem[Shi et~al.(2023)Shi, Gai, Darrell, and Wang]{shi2023refocusing}
Shi, B., Gai, S., Darrell, T., and Wang, X.
\newblock Refocusing is key to transfer learning.
\newblock \emph{arXiv preprint arXiv:2305.15542}, 2023.

\bibitem[Subramani et~al.(2022)Subramani, Suresh, and Peters]{subramani2022extracting}
Subramani, N., Suresh, N., and Peters, M.~E.
\newblock Extracting latent steering vectors from pretrained language models.
\newblock \emph{arXiv preprint arXiv:2205.05124}, 2022.

\bibitem[Torralba \& Efros(2011)Torralba and Efros]{5995347}
Torralba, A. and Efros, A.~A.
\newblock Unbiased look at dataset bias.
\newblock In \emph{CVPR 2011}, pp.\  1521--1528, 2011.
\newblock \doi{10.1109/CVPR.2011.5995347}.

\bibitem[Vaswani et~al.(2017)Vaswani, Shazeer, Parmar, Uszkoreit, Jones, Gomez, Kaiser, and Polosukhin]{vaswani2017attention}
Vaswani, A., Shazeer, N., Parmar, N., Uszkoreit, J., Jones, L., Gomez, A.~N., Kaiser, {\L}., and Polosukhin, I.
\newblock Attention is all you need.
\newblock \emph{Advances in neural information processing systems}, 30, 2017.

\bibitem[Wang et~al.(2023)Wang, Li, Chen, Zhu, Lin, Cao, Liu, Liu, and Sui]{wang2023large}
Wang, P., Li, L., Chen, L., Zhu, D., Lin, B., Cao, Y., Liu, Q., Liu, T., and Sui, Z.
\newblock Large language models are not fair evaluators.
\newblock \emph{arXiv preprint arXiv:2305.17926}, 2023.

\bibitem[Wolf et~al.(2020)Wolf, Debut, Sanh, Chaumond, Delangue, Moi, Cistac, Rault, Louf, Funtowicz, Davison, Shleifer, von Platen, Ma, Jernite, Plu, Xu, Le~Scao, Gugger, Drame, Lhoest, and Rush]{wolf-etal-2020-transformers}
Wolf, T., Debut, L., Sanh, V., Chaumond, J., Delangue, C., Moi, A., Cistac, P., Rault, T., Louf, R., Funtowicz, M., Davison, J., Shleifer, S., von Platen, P., Ma, C., Jernite, Y., Plu, J., Xu, C., Le~Scao, T., Gugger, S., Drame, M., Lhoest, Q., and Rush, A.
\newblock Transformers: State-of-the-art natural language processing.
\newblock In Liu, Q. and Schlangen, D. (eds.), \emph{Proceedings of the 2020 Conference on Empirical Methods in Natural Language Processing: System Demonstrations}, pp.\  38--45, Online, October 2020. Association for Computational Linguistics.
\newblock \doi{10.18653/v1/2020.emnlp-demos.6}.
\newblock URL \url{https://aclanthology.org/2020.emnlp-demos.6}.

\bibitem[Yang et~al.(2023)Yang, Nushi, Palangi, and Mirzasoleiman]{yang2023mitigating}
Yang, Y., Nushi, B., Palangi, H., and Mirzasoleiman, B.
\newblock Mitigating spurious correlations in multi-modal models during fine-tuning.
\newblock \emph{arXiv preprint arXiv:2304.03916}, 2023.

\bibitem[Zhang \& R{\'e}(2022)Zhang and R{\'e}]{zhang2022contrastive}
Zhang, M. and R{\'e}, C.
\newblock Contrastive adapters for foundation model group robustness.
\newblock \emph{Advances in Neural Information Processing Systems}, 35:\penalty0 21682--21697, 2022.

\bibitem[Zhang et~al.(2023)Zhang, Singh, Liu, Liu, Yu, Gao, and Zhao]{zhang2023tell}
Zhang, Q., Singh, C., Liu, L., Liu, X., Yu, B., Gao, J., and Zhao, T.
\newblock Tell your model where to attend: Post-hoc attention steering for llms.
\newblock \emph{arXiv preprint arXiv:2311.02262}, 2023.

\bibitem[Zhao et~al.(2021)Zhao, Wallace, Feng, Klein, and Singh]{zhao2021calibrate}
Zhao, Z., Wallace, E., Feng, S., Klein, D., and Singh, S.
\newblock Calibrate before use: Improving few-shot performance of language models.
\newblock In \emph{International Conference on Machine Learning}, pp.\  12697--12706. PMLR, 2021.

\bibitem[Zheng et~al.(2023{\natexlab{a}})Zheng, Zhou, Meng, Zhou, and Huang]{zheng2023large}
Zheng, C., Zhou, H., Meng, F., Zhou, J., and Huang, M.
\newblock Large language models are not robust multiple choice selectors.
\newblock \emph{arXiv e-prints}, pp.\  arXiv--2309, 2023{\natexlab{a}}.

\bibitem[Zheng et~al.(2023{\natexlab{b}})Zheng, Chiang, Sheng, Zhuang, Wu, Zhuang, Lin, Li, Li, Xing, et~al.]{zheng2023judging}
Zheng, L., Chiang, W.-L., Sheng, Y., Zhuang, S., Wu, Z., Zhuang, Y., Lin, Z., Li, Z., Li, D., Xing, E., et~al.
\newblock Judging llm-as-a-judge with mt-bench and chatbot arena.
\newblock \emph{arXiv preprint arXiv:2306.05685}, 2023{\natexlab{b}}.

\end{thebibliography}
\bibliographystyle{icml2024}

\newpage
\appendix
\onecolumn
\section{Glossary}
\label{appendix:glossary}
The glossary is given in Table~\ref{table:glossary}.
\begin{table*}[h]
\centering
\begin{tabular}{l l}
\toprule
Symbol & Definition \\
\midrule
$q, q'$ & question string, same question with modified option positions \\
$\mathbbm{r}$ & bias rule set \\
$r_j$ & $j$th rule in bias rule set \\
$s_j(q)$ & demonstration of $j$th bias rule from $q$\\
$\mathcal{S}_j$ & demonstration set of $j$th rule ($r_j$) \\
$x$ & input vector in model layers \\
$l$ & model layer \\
$h$ & model single attention head \\
$\textbf{H}$ & matrix of stacked attention head activation values \\
$\theta^x_{h,l}$ & attention head $h$ of layer $l$ activation value of input $x$ \\
$\textbf{v}^j_{h,l}$ & $j$th rule bias direction at head $h$ of layer $l$ \\
$\tilde{\textbf{v}}_{h,l}$ & combined bias direction at head $h$ of layer $l$  \\
$\alpha$ &  hyperparameter to control intervention strength \\
$K$ & hyperparameter: how many attention heads to intervene \\
$N$ & number of samples in dataset \\
\toprule
\end{tabular}
\caption{
	Glossary of variables and symbols used in this paper.
}
\label{table:glossary}
\end{table*}

\section{Limitations}
\label{appendix:limitations}
Our work comes with several limitations that merit attention. Firstly, the accurate identification of bias directions by $\SYSNAME$ depends on the existence of directions in activation space that effectively summarize bias. This necessitates a reasonable separation between usable knowledge and bias within the model's learned representation. The precise circumstances under which these conditions hold remain unclear.

Secondly, we acknowledge the absence of \textit{theoretical} characterizations for hyperparameters and the required number of unlabeled samples. While addressing this theoretical gap would enhance our understanding, there are also practical improvements to be made in $\SYSNAME$. One avenue is the extension of our method to accommodate \textit{non-enumerable} bias. For instance, exploring latent directions associated with harmful text generation, misinformation, or other forms of bias, utilizing principles derived from our work.


\section{Constructing Bias Demonstrations For MCQ Dataset}
\label{appendix:mcq_demonstration}
This section details the bias rule set $\mathbbm{r}$ and demonstration set $\mathcal{S}$ construction for MCQ datasets.
\paragraph{Enumerating Bias Rules.}
Similar to yes/no questions case we demonstrated in Section \ref{sec:method}, MCQ questions with $m$ options have $m$ rules. For example, when we have 3 options: (A/B/C), our rule set $\mathbbm{r}$ items are:
\begin{align*}
r_1 = \texttt{Always choose (A)} \\
r_2 = \texttt{Always choose (B)} \\
r_3 = \texttt{Always choose (C)}
\end{align*}
\paragraph{Constructing Bias Demonstrations.} Unlike yes/no questions, where there are only 2 possible option orderings (``yes/no" and ``no/yes"), the number of orderings for MCQ questions grows in factorial order with the number of presented options $m$ (number of possible orders $= m!$). While our method requires only the model's output to identify bias direction (no decoding necessary), conducting $m!$ forward passes to cover all permutations is computationally intractable. Therefore, we adopt a practical alternative: \textbf{cyclic permutation}, reducing the permutations from $m!$ to a manageable $m$. For example, if our question $q = $ ``Which city is located in Asia? (A) London (B) Chennai (C) Buenos Aires " with three options (A/B/C), we have the following permutations: 
\begin{align*}
    q' = \text{Which city is located in Asia? (A) Buenos Aires (B) London (C) Chennai}\\
    q'' = \text{Which city is located in Asia? (A) Chennai (B) Buenos Aires (C) London}
\end{align*}
The demonstration set $\mathcal{S}_1 = \left \{ s_1(q),  s_1(q'), s_1(q'')\right \}$ is
\begin{align*}
    & s_1(q) = q + \text{``Answer: (A) London"} \\
    & s_1(q') = q' + \text{``Answer: (A) Buenos Aires"} \\
    & s_1(q'') = q'' + \text{``Answer: (A) Chennai"} \\
\end{align*}
. Similarly, for $\mathcal{S}_2 = \left \{  s_2(q),  s_2(q'), s_2(q'')\right \}$ and $\mathcal{S}^3 = \left \{  s_3(q),  s_3(q'), s_3(q'') \right \}$, we append $q, q', q''$ with the answers (B) and (C) respectively.

\section{Dataset Statistics and Setup}
\label{appendix:datasets}
\subsection{Option bias datasets}
Table \ref{tab:option_bias_data} shows dataset statistics for option bias.
\begin{table}[h]
     \begin{center}
     \begin{tabular}{ccccc}
     \toprule
      Dataset & Type & \# Options & \# Test samples & \# Samples for finding direction  \\ 
        \midrule
        \multirow{4}{*}{ScienceQA} & \multirow{4}{*}{MCQ} & 2 & 2228 & \multirow{4}{*}{1000} \\
        & & 3 & 971 & \\
        & & 4 & 1004 & \\
        & & 5 & 38 & \\
        \midrule
        MME & yes/no & 2 & 1,542 & 100 \\
        \midrule
        VGR & yes/no & 2 & 9,576 & 1000 \\
      \bottomrule
      \end{tabular}
      \caption{Dataset statistics for option bias}
      \label{tab:option_bias_data}
      \end{center}
\end{table}
      
Originally, the VGR dataset was a 2 choice options dataset. We are an image and 2 choices: one option is the correct caption of the given image (e.g., ``The cow is eating the grass"), and the other is a false caption (e.g., ``The grass is eating the cow''). We convert this dataset into a yes/no question by turning each caption into 2 questions (e.g., ``Is this the correct caption for the image? answer with a yes or no. The cow is eating the grass"). For ScienceQA, we use default test samples from the original dataset. For MME, we randomly sample 100 samples to find bias direction and use the rest for evaluation. For VGR, we randomly split the dataset 80:20 train/validation:test split, and randomly sample 1000 samples from the train split to find bias direction.

For ScienceQA, since the number of test samples varies significantly between each \# of options, we report the weighted average accuracy (by the number of sample). Other datasets are pretty balance so we report the non-weighted accuracy.

\section{Prompt Details}
\label{appendix:prompt}

We use each model's default system prompts for formatting, detailed as follows:

\textbf{LLaVA}

\lstset{style=mystyle}
\begin{lstlisting}
prompt = A chat between a curious human and an artificial intelligence assistant.  
The assistant gives helpful, detailed, and polite answers to the human's questions.
Human: [QUESTION]
Assistant: [ANSWER]
\end{lstlisting}
For inference, we leave the part after ``Assistant:" empty for the model's answers. For collecting activation values (section \ref{sec:method}), we append the answers based on the constructed demonstration set after ``Assistant:".

\textbf{IDEFICS}

We use the recommended system prompt from IDEFICS Huggingface \cite{wolf-etal-2020-transformers} page \href{https://huggingface.co/HuggingFaceM4/idefics-9b-instruct}{https://huggingface.co/HuggingFaceM4/idefics-9b-instruct}. 
\lstset{style=mystyle}
\begin{lstlisting}
prompt = [
            [
                f"User: {QUESTION}",
                "<end_of_utterance>",
                f"\nAssistant: {ANSWER}",
            ],
        ]
\end{lstlisting}
For inference, we leave the part after ``Assistant:" empty for the model's answers. For collecting activation values (section \ref{sec:method}), we append the answers based on the constructed demonstration set after ``Assistant:".

\textbf{InstructBLIP}

We follow InstructBLIP usage from Huggingface page \href{https://huggingface.co/Salesforce/instructblip-vicuna-13b}{https://huggingface.co/Salesforce/instructblip-vicuna-13b}. , where there is no system prompt. We only use the question string as it is, followed by the answer.
\lstset{style=mystyle}
\begin{lstlisting}
prompt = "[QUESTION] [ANSWER]"
\end{lstlisting}
Similarly for the previous two cases, we leave the part after the question empty for inference, and fill the answers based on the constructed demonstration set for collecting activation values.


\section{Implementation Details}
\subsection{Compute details}
There is no Transformer model training or fine-tuning conducted in this paper's experiments. We use 8 Test V100 GPUs for hyperparameter tuning and evaluation. 
\subsection{$\SYSNAME$ implementation}
We use IDEFICS and InstructBLIP models from HuggingFace \cite{wolf-etal-2020-transformers} and LLaVA from the author's repository \cite{liu2023llava}. We provide pseudocode for collecting model activation values in Algorithm table \ref{alg:set_demo}.
\begin{algorithm}[H]
        \caption{Pseudocode for collecting activation values $\mathbbm{H}$} \label{alg:set_demo}
        \begin{flushleft}
	\begin{algorithmic}[1]
		\STATE \textbf{Parameters:}
            Demonstration sets $\mathcal{S}_1, \ldots \mathcal{S}_m$, model $T$, attention head index $h$, layer index $l$, question sets $\{q_1,\ldots q_N \}, \{q'_1,\ldots q'_N \}, ...$ \\
            \FOR{$i \in \left \{1,\ldots,N \right \}$}
            \FOR{$j \in \left \{1,\ldots,m \right \}$}
            \STATE $\textbf{H}^j$ = []
            \FOR{$\left \{ s_j(q_i)\right \} \in \mathcal{S}^j$}
                \STATE $\theta_{h,l}^{s_j(q_i)}$ = T($s_j(q_i)$)[“hidden states”][-1][$l,h$]
                \STATE $\textbf{H}^j$.append($\theta_{h,l}^{s_j(q_i)}$)
             \ENDFOR 
             \STATE $\textbf{H}^j = \text{np.vstack}(\textbf{H}^j)$
            \ENDFOR  
            \ENDFOR
            \STATE \textbf{Returns:}  $\textbf{H}^1,\ldots \textbf{H}^m$.
            
	\end{algorithmic} 
        \end{flushleft}
\end{algorithm}

\subsection{Baselines implementation}
This section presents implementation details for baseline methods.
\paragraph{Option bias}
Vanilla inference is done with prompts detailed in Appendix \ref{appendix:prompt}. ITI \cite{li2023inferencetime} code is adapted from the author's original repository \href{https://github.com/likenneth/honest_llama}{https://github.com/likenneth/honest\_llama}. The 100 and 500 samples are randomly sampled from the datasets training splits.

\paragraph{Stereotypical bias} Vanilla inference is done with prompts detailed in Appendix \ref{appendix:prompt}. Prompting baseline uses the following prompt prepended to each question: ``Do not stereotype."

\subsection{Hyperparameter search}
We perform hyperparameter search for both $\SYSNAME$ and ITI. We list the hyperparameter search space in Table \ref{tab:hyperparams}

\begin{table}[ht!]
     \begin{center}
     \begin{tabular}{ccc}
     \toprule
      Method & Intervention strength $\alpha$ & number of heads $K$ \\ 
    \midrule
    ITI & $\left \{1,5,10,15,20,25,30,40,50 \right \}$ & $\left \{1,10,20,30,40,50,100 \right \}$ \\
    \midrule
    $\SYSNAME$ & $\left \{0.1,0.5,1,2,5,10,15,20,25 \right \}$  & $\left \{10,30,50,100,200,500 \right \}$  \\
      \bottomrule
      \end{tabular}
      \caption{Hyperparameter search space}
      \label{tab:hyperparams}
      \end{center}
      \end{table}
We follow the initial hyperparameter space for for as suggested in the original paper \cite{li2023inferencetime}. We try a larger number of $K$ in $\SYSNAME$ because we use $l_2$ normalization post-intervention (Section \ref{sec:method}). The best hyperparameter is chosen based on the best performance on the validation set (minival split for ScienceQA).

\label{appendix:implementation}

\section{Supplementary Results}
\label{appendix:supp_results}
We present exhaustive, non averaged results in this section.

\begin{figure}[ht!]
    \centering 
\begin{subfigure}{0.23\textwidth}
  \includegraphics[width=\linewidth]{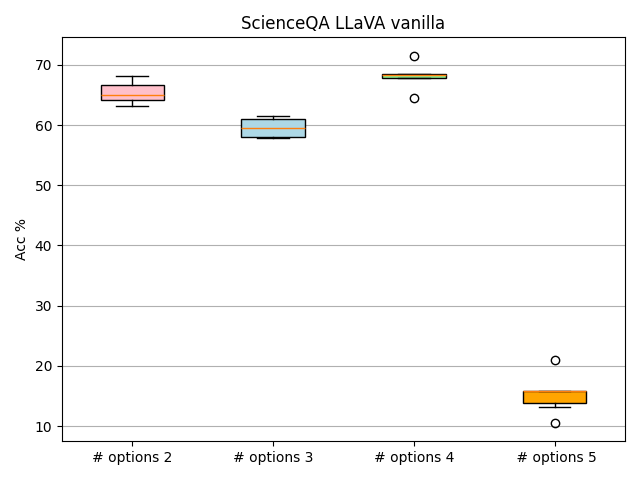}
\end{subfigure} 
\begin{subfigure}{0.23\textwidth}
  \includegraphics[width=\linewidth]{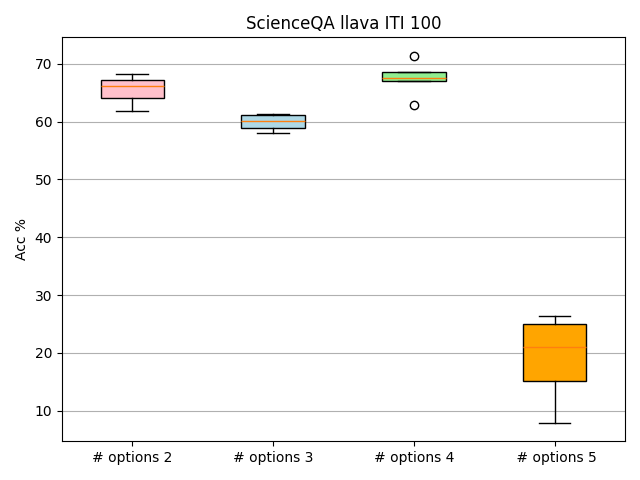}
\end{subfigure}
\begin{subfigure}{0.23\textwidth}
  \includegraphics[width=\linewidth]{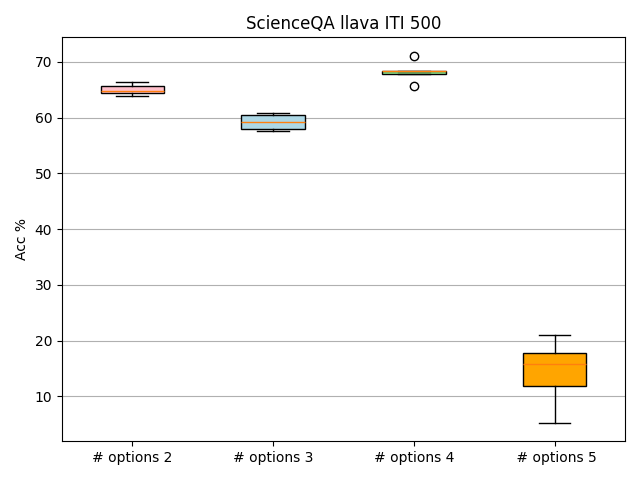}
\end{subfigure}
\begin{subfigure}{0.23\textwidth}
  \includegraphics[width=\linewidth]{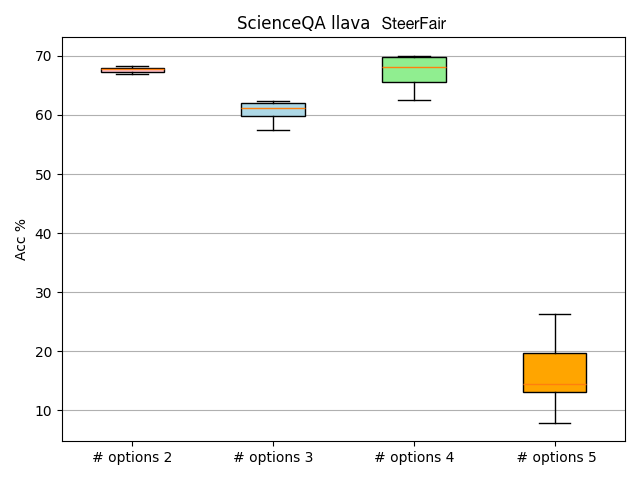}
\end{subfigure}

\medskip
\begin{subfigure}{0.23\textwidth}
  \includegraphics[width=\linewidth]{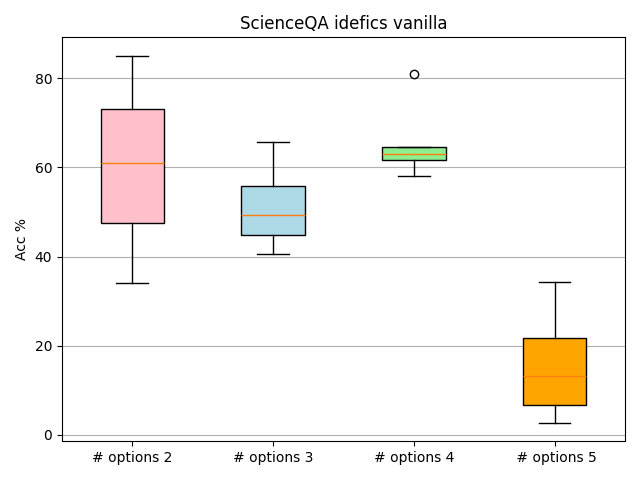}
\end{subfigure}
\begin{subfigure}{0.23\textwidth}
  \includegraphics[width=\linewidth]{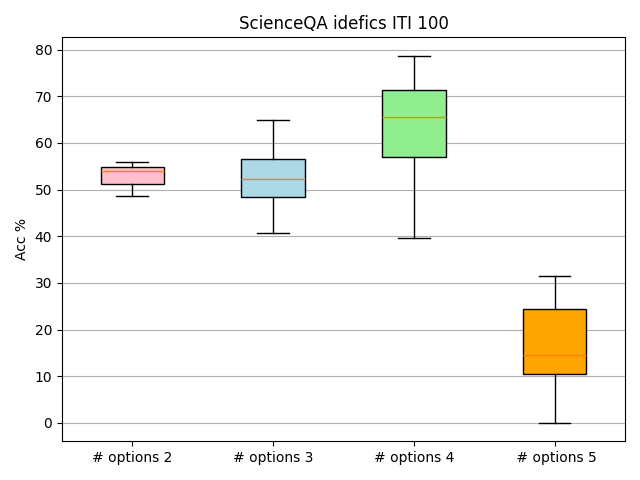}
\end{subfigure}
\begin{subfigure}{0.23\textwidth}
  \includegraphics[width=\linewidth]{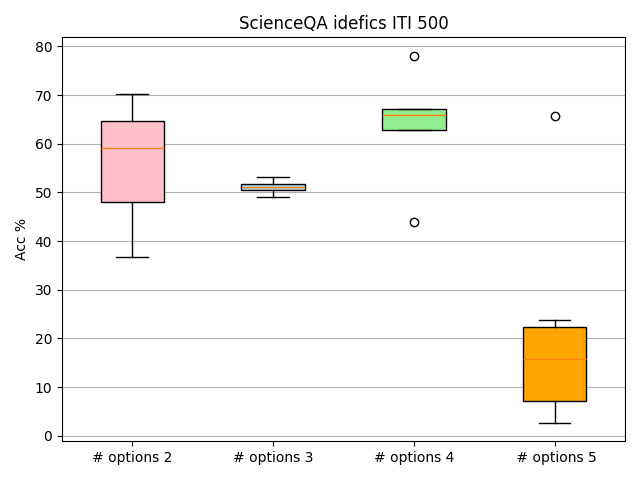}
\end{subfigure}
\begin{subfigure}{0.23\textwidth}
  \includegraphics[width=\linewidth]{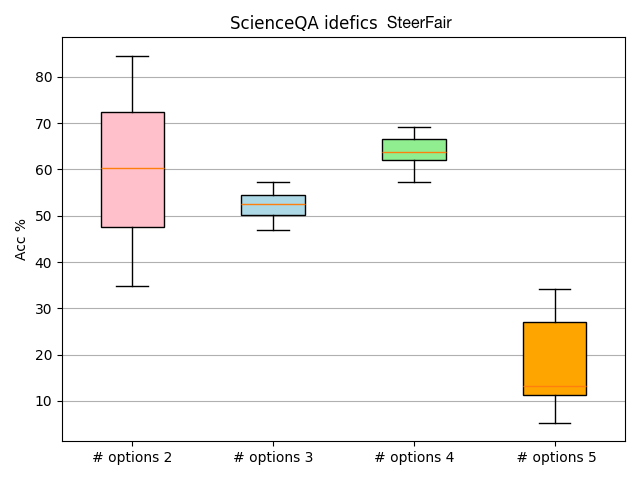}
\end{subfigure}

\medskip
\begin{subfigure}{0.23\textwidth}
  \includegraphics[width=\linewidth]{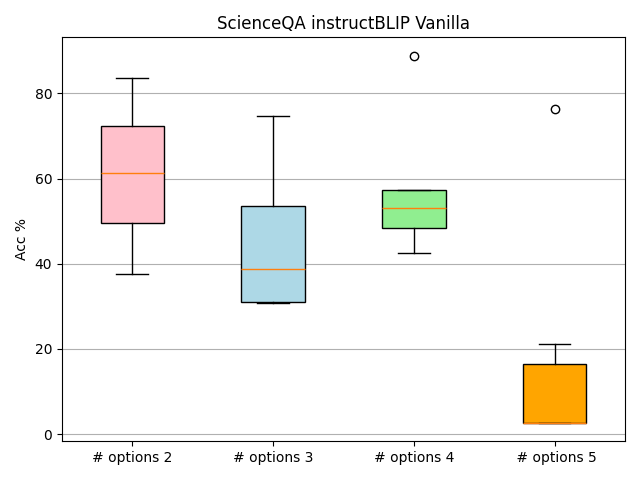}
\end{subfigure}
\begin{subfigure}{0.23\textwidth}
  \includegraphics[width=\linewidth]{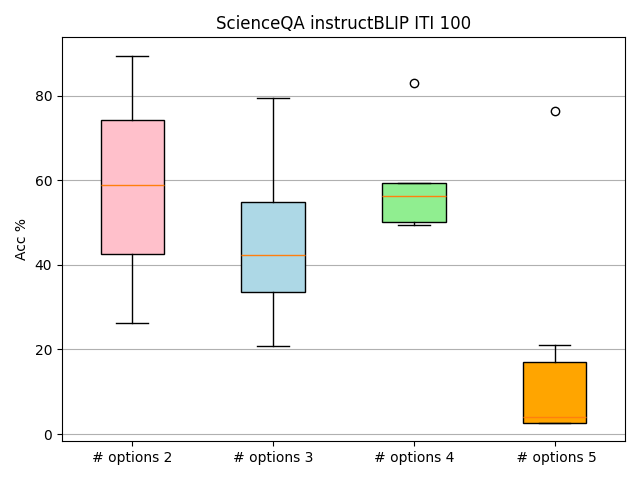}
\end{subfigure}
\begin{subfigure}{0.23\textwidth}
  \includegraphics[width=\linewidth]{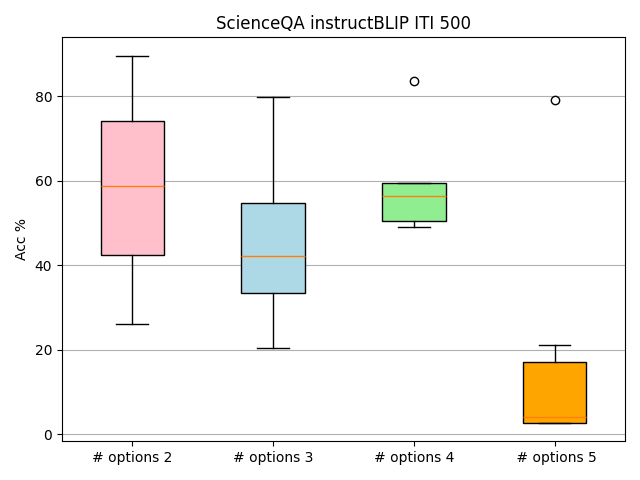}
\end{subfigure}
\begin{subfigure}{0.23\textwidth}
  \includegraphics[width=\linewidth]{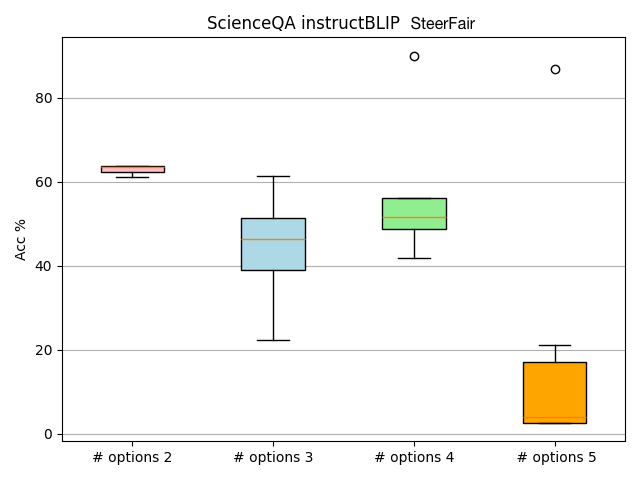}
\end{subfigure}
\caption{ScienceQA results}
\label{fig:sqa_result}
\end{figure}

\begin{figure}[ht!]
    \centering 
\begin{subfigure}{0.23\textwidth}
  \includegraphics[width=\linewidth]{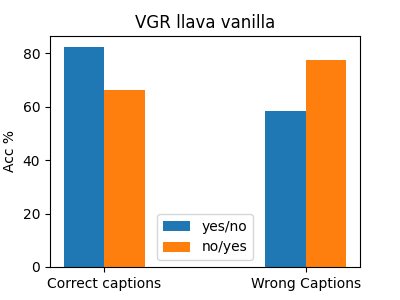}
\end{subfigure} 
\begin{subfigure}{0.23\textwidth}
  \includegraphics[width=\linewidth]{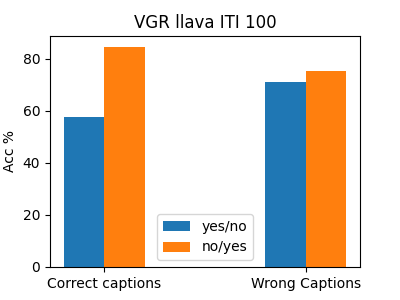}
\end{subfigure}
\begin{subfigure}{0.23\textwidth}
  \includegraphics[width=\linewidth]{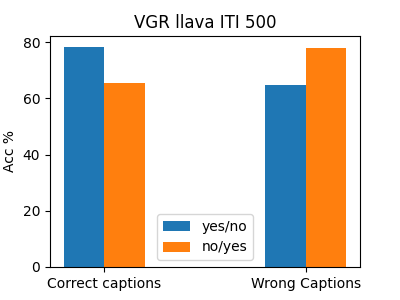}
\end{subfigure}
\begin{subfigure}{0.23\textwidth}
  \includegraphics[width=\linewidth]{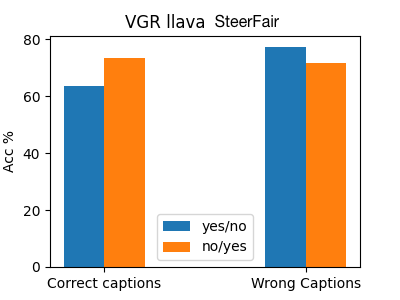}
\end{subfigure}

\medskip
\begin{subfigure}{0.23\textwidth}
  \includegraphics[width=\linewidth]{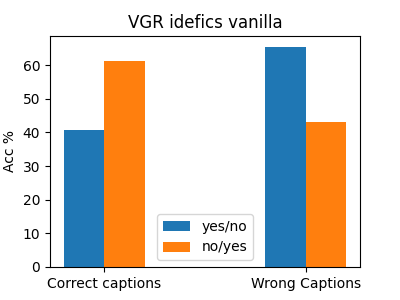}
\end{subfigure}
\begin{subfigure}{0.23\textwidth}
  \includegraphics[width=\linewidth]{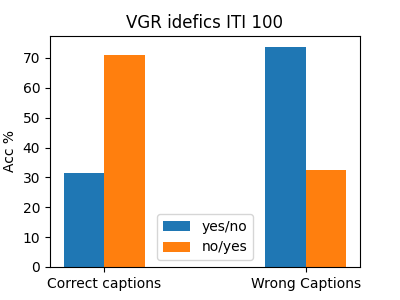}
\end{subfigure}
\begin{subfigure}{0.23\textwidth}
  \includegraphics[width=\linewidth]{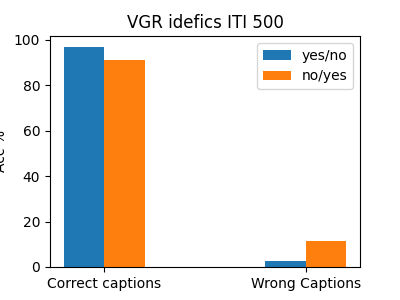}
\end{subfigure}
\begin{subfigure}{0.23\textwidth}
  \includegraphics[width=\linewidth]{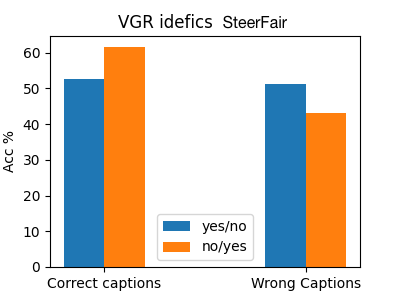}
\end{subfigure}

\medskip
\begin{subfigure}{0.23\textwidth}
  \includegraphics[width=\linewidth]{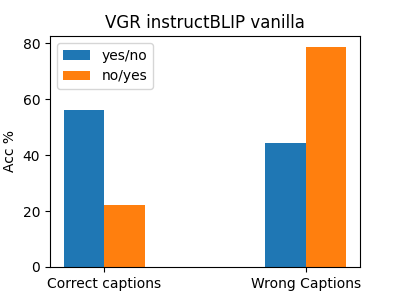}
\end{subfigure}
\begin{subfigure}{0.23\textwidth}
  \includegraphics[width=\linewidth]{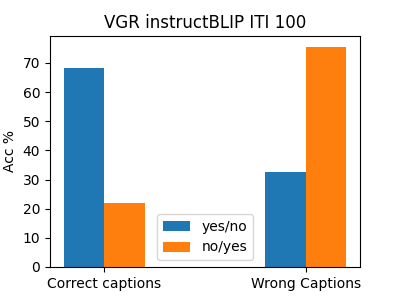}
\end{subfigure}
\begin{subfigure}{0.23\textwidth}
  \includegraphics[width=\linewidth]{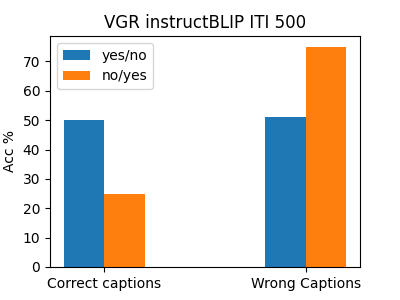}
\end{subfigure}
\begin{subfigure}{0.23\textwidth}
  \includegraphics[width=\linewidth]{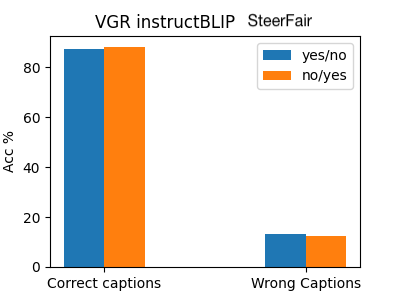}
\end{subfigure}
\caption{VGR results}
\label{fig:vgr_result}
\end{figure}

\begin{figure}[ht!]
    \centering 
\begin{subfigure}{.8\textwidth}
  \includegraphics[width=\linewidth]{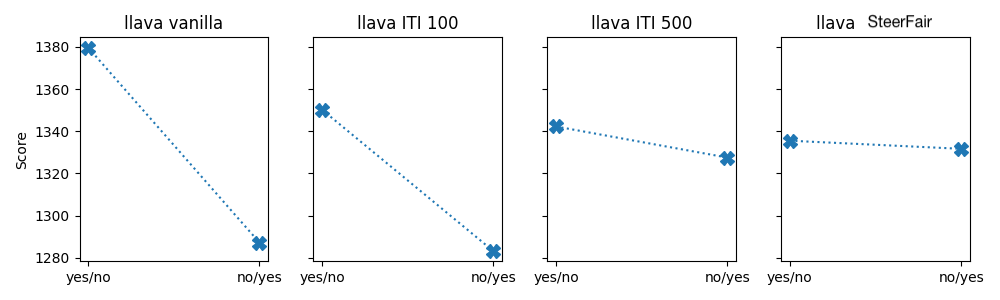}
\end{subfigure} 
\medskip
\begin{subfigure}{.8\textwidth}
  \includegraphics[width=\linewidth]{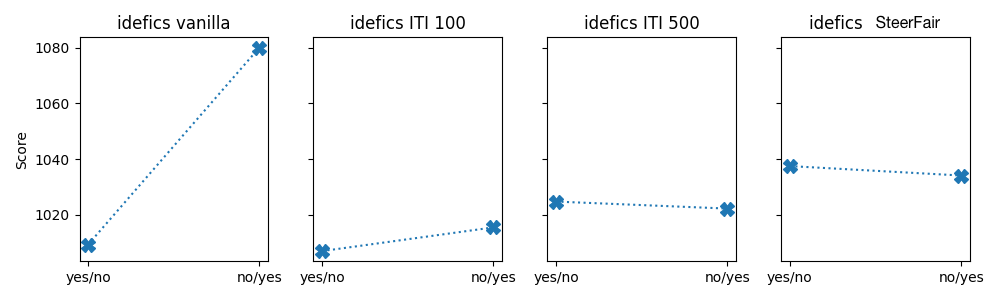}
\end{subfigure}
\medskip
\begin{subfigure}{.8\textwidth}
  \includegraphics[width=\linewidth]{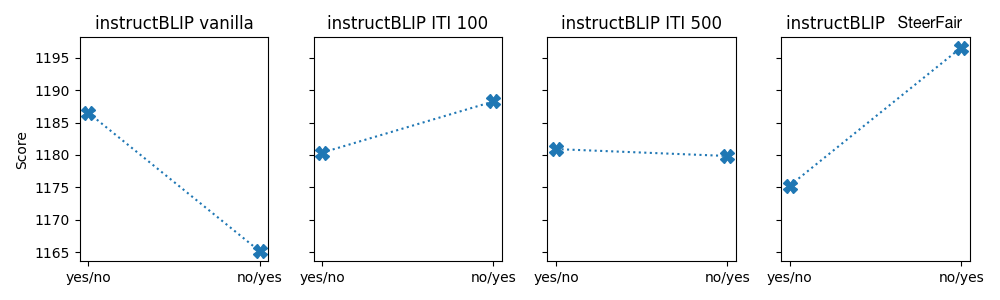}
\end{subfigure}
\caption{MME results}
\label{fig:mme_result}
\end{figure}



\end{document}